\pdfoutput=1

\documentclass[11pt]{article}

\usepackage[final]{acl}

\usepackage{times}
\usepackage{latexsym}

\usepackage[T1]{fontenc}

\usepackage[utf8]{inputenc}

\usepackage{microtype}

\usepackage{inconsolata}
\usepackage{flushend}
\usepackage[a-1b]{pdfx}   
\usepackage{balance}

\usepackage{color}
\usepackage{tcolorbox}
\usepackage[labelfont=bf]{caption}
\usepackage{subcaption}
\usepackage{hyperref}
\usepackage{makecell}
\usepackage{fixltx2e}
\usepackage{rotating,multirow}
\usepackage{etoolbox,xspace}
\usepackage{algorithmicx}
\usepackage{algorithm}
\usepackage{fontawesome}
\usepackage{algpseudocode}
\usepackage{wrapfig}
\usepackage{scalerel,amssymb}
\usepackage{booktabs}
\usepackage{multirow}
\usepackage{siunitx}
\usepackage{amsmath}

\algtext*{EndWhile}
\algtext*{EndIf}
\algtext*{EndFunction}
\algtext*{EndFor}
\algrenewcommand\algorithmicrequire{\textbf{Input:}}
\algrenewcommand\algorithmicensure{\textbf{Output:}}

\usepackage{amsthm}

\usepackage{tabularx}
\usepackage{ragged2e}  
\usepackage{booktabs}  
\usepackage{textcomp}
\usepackage{url}
\usepackage{comment}
\usepackage{enumitem}
\usepackage[simplified]{pgf-umlcd}
\usepackage{tikz}

\newcolumntype{Y}{>{\RaggedRight\arraybackslash}X}

\newtheorem{definition}{Definition}

\newtcolorbox{defbox}{
  colback=orange!10,
  colframe=orange!20,
  arc=2mm, 
  fonttitle=\bfseries,
  boxrule=0mm,
  boxsep=1mm,
  left=0mm,
  right=0mm,
  top=0mm,
  bottom=0mm
}

\newcounter{example}
\renewcommand{\theexample}{\arabic{example}}

\newtcolorbox{examplebox}{
  colback=blue!10,
  colframe=blue!20,
  arc=2mm, 
  fonttitle=\bfseries,
  boxrule=0mm,
  boxsep=1mm,
  left=0mm,
  right=0mm,
  top=0mm,
  bottom=0mm
}

\newenvironment{example}{
  \refstepcounter{example}
  \begin{examplebox}{}
    \noindent {\bf Example \theexample:} 
  }
  {\end{examplebox}} 

\newcommand{\squishlist}{
 \begin{list}{$\bullet$}
  { \setlength{\itemsep}{0pt}
     \setlength{\parsep}{1pt}
     \setlength{\topsep}{1pt}
     \setlength{\partopsep}{0pt}
     \setlength{\leftmargin}{1em}
     \setlength{\labelwidth}{1em}
     \setlength{\labelsep}{0.5em} } }

\newcommand{\squishend}{
  \end{list}
}

\definecolor{americanrose}{rgb}{1.0, 0.01, 0.24}
\definecolor{airforceblue}{rgb}{0.36, 0.54, 0.66}
\definecolor{ao(english)}{rgb}{0.0, 0.5, 0.0}
\definecolor{ao}{rgb}{0.0, 0.0, 1.0}

\newcommand{\eat}[1]{}

\usepackage{xspace}
\newcommand{\stitle}[1]{\vspace{2mm}\noindent{\bf #1:}\xspace}
\newcommand{\at}[1]{{\sc \small #1}\xspace}

\newcommand{\similarity}{\mathcal{S}_{im}}

\newcommand{\Gee}{\mathcal{G}}
\newcommand{\gee}{\mathbf{g}}

\newcommand{\dist}{\xi}

\newcommand{\floor}[1]{\left \lfloor #1 \right \rfloor}

\newcommand{\stereo}{\textsf{StereoSet}\xspace}
\newcommand{\wino}{\textsf{WinoBias}\xspace}
\newcommand{\forbes}{\textsf{Forbes 2022 Billionaire}\xspace}
\newcommand{\student}{\textsf{Students}\xspace}

\newcommand{\system}{\textsc{Requal-lm}\xspace}
\newcommand{\blender}{\textsc{Pair-Ranker}\xspace}
\newcommand{\gptdebias}{\textsc{Debiased-GPT}\xspace}
\newcommand{\first}{\textsc{First-Response}\xspace}

\newcommand{\wo}{{\sc\small Weighted Output}\xspace}
\newcommand{\uo}{{\sc\small Unweighted Output}\xspace}
\newcommand{\mbo}{{\sc\small Min-bias Output}\xspace}

\title{\system: \underline{R}eliability and \underline{Equ}ity through \underline{A}ggregation \\in \underline{L}arge \underline{L}anguage \underline{M}odels\thanks{This work was supported in part by NSF 2107290.}}

\author{Sana Ebrahimi \\
  University of Illinois Chicago \\
  \texttt{sebrah7@uic.edu} \\\And
  Nima Shahbazi \\
  University of Illinois Chicago \\
  \texttt{nshahb3@uic.edu} \\\And
  Abolfazl Asudeh \\
  University of Illinois Chicago \\
  \texttt{asudeh@uic.edu} \\}

\begin{document}
\maketitle
\begin{abstract}
The extensive scope of large language models (LLMs) across various domains underscores the critical importance of responsibility in their application, beyond natural language processing.
In particular, the randomized nature of LLMs, coupled with inherent biases and historical stereotypes in data, 
raises critical concerns regarding reliability and equity.
Addressing these challenges are necessary before using LLMs for applications with societal impact.

Towards addressing this gap, we introduce \system, a novel method for finding reliable and equitable LLM outputs through aggregation.
Specifically, we develop a Monte-carlo method based on repeated sampling to find a reliable output close to the mean of the underlying distribution of possible outputs.
We formally define the terms such as reliability and bias, and design an equity-aware aggregation to minimize harmful bias while finding a highly reliable output.
\system does not require specialized hardware, does not impose a significant computing load, and uses LLMs as a black-box. 
This design choice enables seamless scalability alongside the rapid advancement of LLM technologies. Our system does not require retraining the LLMs, which makes it deployment-ready and easy to adapt.

Our comprehensive experiments using various tasks and datasets demonstrate that \system effectively mitigates bias and selects a more equitable response, specifically the outputs that properly represents minority groups.
\end{abstract}
\section{Introduction}
In the ever-evolving realm of advanced technologies, Large Language Models (LLMs) have quickly emerged as versatile tools, extending their influence far beyond the boundaries of natural language processing (NLP).
Many of the traditionally challenging tasks with decades of research in various fields of computer science are finding more effective resolutions with the help of LLMs. Let us consider Example~\ref{ex-1} as a motivating example for subset selection using LLM.

\begin{example}\label{ex-1}
    {\bf (Part 1)}
    {\it
       Selecting a subset of candidates from a pool, based on a set of criteria is common across multiple applications ranging from journalism, to college admissions and job hiring.
       For example, consider the HR department of a sales company who wants to select a set of candidates for the performance award based on multiple criteria such as \at{sales} and \at{customer-satisfaction}.
       Passing the performance information of the employees, they can ask the LLM to select the candidate set.
    }
\end{example}


LLMs are sequential randomized approaches based on estimations learned from large textual datasets. In particular, based on the prompt and the sequence of tokens generated so far, each word (token) in the dictionary is assigned a probability. Then, the next token is generated probabilistically (proportional to the probabilities of the top-k or top-p\%) using the parameter temperature.
Consequently, the output may vary when the LLM is queried again.
As a result, a valid concern, particularly for a decision maker, is {\em whether they should rely on the LLM's output} for taking action.
In settings similar to Example~\ref{ex-1}, the reliability question is further significant, since a method to combine the performance criteria has not been specified, while small changes in the combination details may significantly change the output~\cite{guan2019mithraranking}. 

Another challenge that makes a single query to the LLMs unreliable arises for the {\em symmetric} settings, where the ordering between the input does not matter, i.e., shuffling the input should not impact the output.
For instance, in Example~\ref{ex-1} the ordering based on which the employees are passed to the LLM should not impact the output.
Conversely, LLMs receive an input as a (ordered) sequence. As a result, as it was observed in~\cite{gao2023double}, the output of the LLMs for symmetric problems vary when the input is shuffled.
We also observed the same behavior in our experiments on a subset selection task, where the entities that are placed at the beginning of the list had a higher chance of being returned as the output.

To resolve these issues
we introduce \system that, instead of relying on a single query to an LLM, follows a Monte-carlo method~\cite{hammersley2013monte} based on repeated sampling.
Particularly, viewing each LLM output as a sample from the underlying distribution of possible outputs, it identifies the centroid of a collection of samples as its estimation of the mean of the distribution, and returns the closest output to the centroid as the most reliable one. To further clarify this, let us consider Example~\ref{ex-1} once again.

\begin{examplebox}
    \noindent{\bf Example~\ref{ex-1}: (Part 2)}
    {\it
        Observing the dependency of the LLM output with the input ordering, and to possibly consider various combinations of performance criteria, the HR department does not rely on a single output of the LLM. Instead \system enables issuing multiple queries to the LLM, each time shuffling the list of the employees. 
        It then returns the ``closest-to-centroid'' of the obtained samples as the most reliable output. 
    }
\end{examplebox}

While being effective in practice, data-driven technologies have been heavily criticized for machine bias~\cite{angwin2022machine}, and LLMs are not an exception when it comes to bias.
As a result, another valid concern when using LLMs for decision making is neutrality: to ensure that impact of historical biases and stereotypes are minimized and that values such as diversity are promoted.

\begin{examplebox}
    \noindent{\bf Example~\ref{ex-1}: (Part 3)}
    {\it
        The HR department would likes to maximize diversity in the selected awardees. In particular, they would like to prevent selecting a male-only list of employees.
        \system allows specifying two or more demographic groups and it minimizes the output bias (measured as the cosine-similarity difference of its output's embedding with different groups' representations).
    }
\end{examplebox}

{
LLMs are among the fast-growing technologies, with new and advanced versions regularly emerging, while many of these systems are ``black-box''.
Our system design is not dependent on a specific LLM, which makes it \underline{\em a ready-to-apply wrapper} that works on top of \underline{\em any} of the current and future closed-source and open-source LLMs.
\system does not require pre-training or fine-tuning, is task-agnostic, and can handle non-binary demographic groups.
}

In the following, first in \S~\ref{sec:problem} we carefully discuss the problem setting, introduce notations, and formally define terms such as reliability and bias.
Next, in \S~\ref{sec:ovreview} we review the architecture of \system, and develop our methodology for finding an equitable centroid and return the closest output to it, the one that is both equitable and reliable.
The experimental evaluations, related work, and the discussions of the benefits and limitations of \system are provided in \S~\ref{sec:exp}, \S~\ref{sec:related}, \S~\ref{sec:benefits}, and \S~\ref{sec:limitations}, respectively.

\section{Preliminaries}\label{sec:problem}


\noindent -- (Input) {\em Task:} We consider a task, such as subset selection, sentence completion, assembling a team of experts, etc., described in form of a prompt: a natural language instruction. 

\noindent -- (Input) {\em Demographic Groups:} We assume the existence of at least one sensitive attribute (e.g., {\tt\small sex}) that specify the demographic groups $\Gee=\{\gee_1, \cdots, \gee_\ell\}$ (e.g., {\tt\small \{male, female\}}). The demographic groups are used to specify the output bias.

\noindent{\em -- LLM:} We assume access to (at least) one LLM, which is used for task answering. The LLM is randomized, i.e., the tokens are sequentially drawn based on the underlying distribution of the (top-k or top-p\%) token-probabilities.
We treat the LLM as a black-box oracle that upon querying generates an {\em output} based on the input prompt.
Treating the LLM as black-box allows the adaptation of \system both for closed-source and open-source LLMs.

\noindent{\em -- Text Embedding:} We rely on an external text embedding model that transforms a text into an embedding vector.
Specifically, given a text $O_i$,
it generates the vector representation $\vec{v}(O_i) = \vec{v}_i = \langle v_1, v_2, \cdots, v_d\rangle$.
Our system, \system, is agnostic to the choice (but limited to the performance) of the embedding model,
and can adapt any state-of-the-art text embedding technique.
Without loss of generality, we use \at{Instructor} -- a method for generating task-specific embeddings in accordance with provided instructions~\cite{Instructor}.

Given two text phrases $O_i$ and $O_j$ and their corresponding embeddings $\vec{v}_i$ and $\vec{v}_j$, the similarity between $O_i$ and $O_j$ is measured as the cosine similarity between their embeddings, i.e., $\similarity(O_i,O_j) = \cos{\angle(\vec{v}_i,\vec{v}_j)}$.
Similarly, the distance between $O_i$ and $O_j$ is defined as $\Delta(O_i,O_j) = 1-\similarity(\vec{v}_i,\vec{v}_j)$.

\begin{defbox}
\begin{definition}[Reliability]\label{def:reliability}
    Given a prompt $I$, let $\mathcal{O}_I$ be the universe of possible-to-generate outputs for $I$. Furthermore, let $\dist$ be the probability distribution of outputs for $I$. That is, $\forall O\in \mathcal{O}_I$, $Pr_\dist(O)$ is the probability that $O$ is generated for 
    $I$. Let $\vec{\mu}_\dist$ be the mean of $\dist$ in the embedding space.
    Then the reliability of an output $O\in\mathcal{O}_I$ is defined as its similarity to $\vec{\mu}_\dist$. That is,
    
    \vspace{-6mm}
    \begin{align*}
        \rho(O) = \similarity(\vec{v}(O),\vec{\mu}_\dist)
    \end{align*}
\end{definition}
\end{defbox}
    
{Let $O\in \mathcal{O}_I$ be an output generated for the prompt $I$ comprising a sequence of $|O|$ tokens $\langle t^O_1, t^O_2, \cdots t^O_{|O|}\rangle$ sequentially generated by the LLM.}
At each iteration $i$, let $Pr(t^O_i)$ be the probability of generating the token $t^O_i$. Then $Pr_\dist(O)$ can be computed as the product of its token probabilities. That is, $Pr_\dist(O) = \prod_{i=1}^{|O|} Pr(t^O_i)$.

\begin{defbox}
\begin{definition}[Bias]\label{def:bias}
    Consider a set of demographic groups $\Gee=\{\gee_1,\cdots,\gee_\ell\}$ and their corresponding vector representation\footnotemark[1] $\{\vec{\gee_1},\cdots,\vec{\gee_\ell}\}$.
    The bias of an output $O$ for a prompt $I$ is computed as the maximum similarity disparity of the demographic groups with $O$. Formally,
    
    \vspace{-6mm}
    \begin{align*}
        \beta(O) = \max_{\gee_i,\gee_j\in \Gee} \big| \similarity(\vec{v}(O), \vec{\gee_i}) -  \similarity(\vec{v}(O), \vec{\gee_j})\big|
    \end{align*}
\end{definition}
\end{defbox}

\footnotetext[1]{Please refer to Appendix~\ref{sec:demovec} for the details of obtaining the vector representations for the demographic groups.}

Bias is sometimes inherent to the task at hand and is not harmful. For example, when the task involves summarizing or rephrasing a paragraph that is particularly written about a specific gender, the resulting output tends to be naturally biased towards that gender. We call this type of output bias as the {\em inevitable bias}.
Formally, we say a bias level $\varepsilon$ is inevitable
if there is no valid output $O\in \mathcal{O}_I$ with a bias less than $\varepsilon$. 
In other words, for any output $O'$ where $\beta(O)<\varepsilon$, we can say $O'\notin \mathcal{O}_I$. 
Therefore, we define the inevitable bias as $\beta_n(I) = \min_{O\in \mathcal{O}} \beta(O)$. 
We consider any bias that is not inevitable, discriminatory.
Harmful stereotypes are in this category. We call this type of output bias as the {\em harmful bias}.
Considering {\em equity} as our objective in this paper, we would like to minimize harmful bias in the outputs. 
The harmful bias of an output can be computed by subtracting its bias from the inevitable bias, i.e., $\beta_h(O) = \beta(O) - \beta_n(I)$.


After defining the terms and notations, we are able to formulate our problem: {\em given a task presented in the form of a prompt $I$, and including the demographic groups $\Gee$, the objective is to identify an output $O\in\mathcal{O}_I$, such that it maximizes $\rho(O)$ and minimizes $\beta_h(O)$}.
\begin{figure*}[!bth]
    \centering
\includegraphics[width=\textwidth]{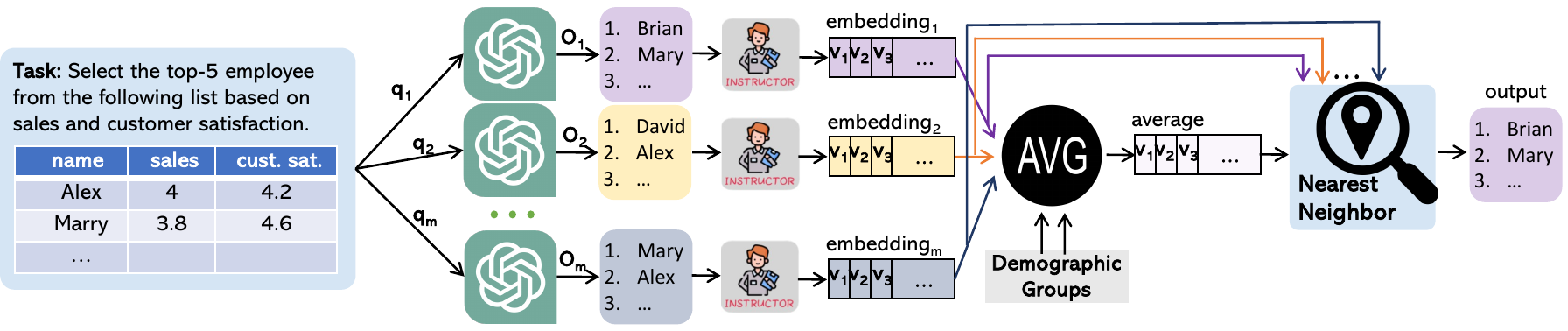}
    \caption{System Architecture of \system.}
    \label{fig:arch}
    \vspace{-5mm}
\end{figure*}

\section{Technical Details}\label{sec:ovreview}
\subsection{Architecture Overview}\label{sec:arch}
Figure~\ref{fig:arch} shows the architecture of \system.
Following the Monte-carlo method described in \S~\ref{sec:approach}, the first step is to obtain a set of iid output samples by issuing $m$ independent queries to the LLM.
 The results are subsequently  fed into the text embedding model, \at{Instructor}, to obtain the vector representations $\{\vec{v}(O_1),\cdots \vec{v}(O_m)\}$.
Next, the vector representations, as well as the vector representations of the demographic groups, pass on to the {\em aggregation function} (referred to as AVG in the figure). The aggregation function generates the vector representation that corresponds to the average of $\vec{v}(O_1)$ to $\vec{v}(O_m)$.
Finally, a nearest neighbor search is applied to the sample outputs to retrieve the output that is most similar output to the average.

\subsection{Methodology}\label{sec:approach}
Our approach for satisfying reliability and equity in LLM outputs is a Monte-carlo method, which relies on repeated sampling and the central limit theorem~\cite{durrett2010probability}.
Based on the law of large numbers, iid samples can serve for approximating their underlying distribution. That is because the expected number of occurrences of each observation is proportional to its probability.

Recall that the outputs for a prompt $I$ are generated based on the probability distribution $\dist$. Particularly, the probability that an output $O\in \mathcal{O}_I$ is sampled is $Pr_\dist(O)$.
Therefore, the expected value of $\vec{v}(O)$ is equal to the mean of $\dist$ in the embedding space, $\vec{\mu}_\dist$.
Now consider a set $\mathbf{O}=\{O_1\cdots, O_m\}$ of iid output samples for the prompt $I$. Let $\vec{v}_c$ be the sample mean of the representation vectors in $\mathbf{O}$. That is,

\vspace{-12mm}
\begin{align}\label{eq:unweighted}
    \vec{v}_c = \frac{1}{m}\sum_{i=1}^m \vec{v}(O_i)
\end{align}
Similarly, let $\vec{\sigma}$ be the standard deviation of the samples.
Following the central limit theorem, $\vec{v}_c$ follows $\mathcal{N}\big(\vec{\mu}_\dist, \frac{\vec{\sigma}}{\sqrt{m}}\big)$, the Normal distribution with the mean $\vec{\mu}_\dist$ and standard deviation $\frac{\vec{\sigma}}{\sqrt{m}}$.
For simplicity, in the rest of the paper, we call $\vec{v}_c$ the {\bf centroid} of the output samples.

\system considers two approaches for specifying the value of $m$: (i) fixed budget and (ii) fixed error.
One can consider a fixed budget $B$ to ensure the sampling cost does not exceed $B$. Specifically, if the cost of each query is $c$, then $m=\floor{\frac{B}{c}}$.
Alternatively, when a flexible budget is available, one can collect enough samples to bound the confidence error for a specific confidence level $\alpha$ (e.g., 95\%). The confidence error $\vec{e}$ guarantees $Pr(|\vec{v}_c - \vec{\mu}_\dist|>\vec{e})\leq 1-\alpha$.
Following the central limit theorem and using the Z-table, the confidence error is computed as $\vec{e} = Z(1-\frac{\alpha}{2})\frac{\vec{\sigma}}{\sqrt{m}}$. 

\subsection{Equity-aware Aggregation}

Using the centroid of sample outputs $\mathbf{O}$ as the estimation of $\vec{\mu}_\dist$, we can estimate the reliability of each output $O\in\mathbf{O}$ as $E\big[\rho(O)\big] = \similarity(\vec{v}(O),\vec{v}_c)$, and identify the output with the maximum expected reliability.

\begin{figure}[!tb]
    \centering
    \includegraphics[width=0.36\textwidth]{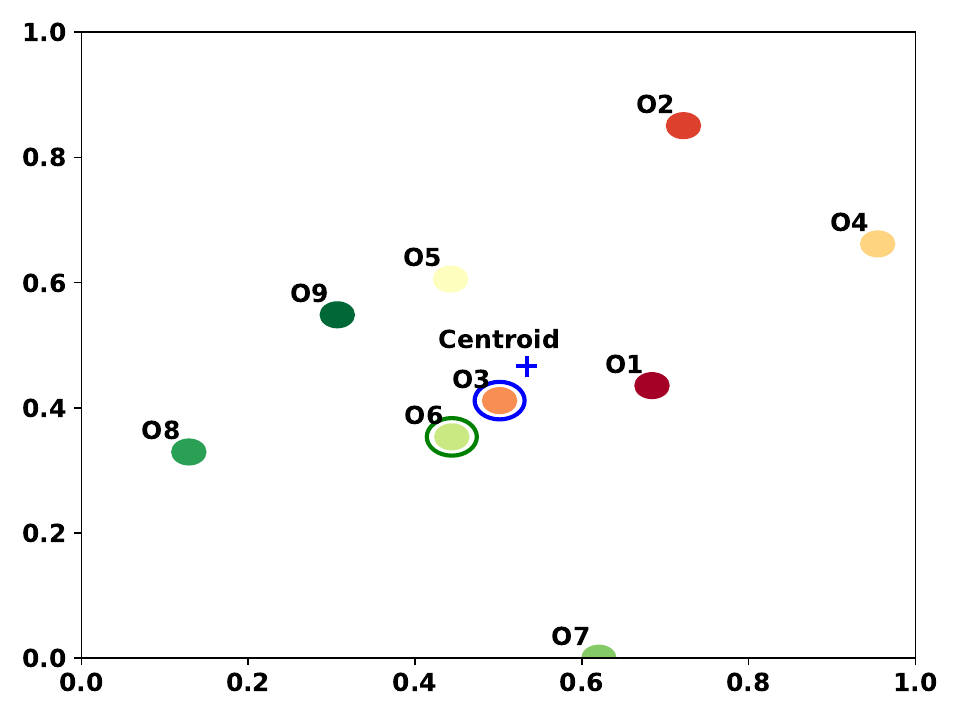}
    \caption{A toy t-SNE visualization of nine output samples, and their centroid. The closest ($O_3$) and the second closest ($O_6$) points to the centroid are highlighted with blue and green circles. The green-to-red color code shows the bias values.}
    \label{fig:t-sne}
\end{figure}

Figure~\ref{fig:t-sne} shows a toy T-SNE visualization of $9$ sample outputs, while their centroid is marked with a plus sign. The distance of the points from the centroid show their expected reliability. In this example, $O_3$ is the most reliable output.
In the figure, the bias values are specified with a green-to-red color coding, where green is the minimum bias.
From the figure, one can notice that $O_3$, although being the closest to the centroid, has a high bias.
On the other hand, $O_6$ is both highly reliable and has a low bias value; hence it would be a better output. In order to achieve both objectives of high reliability and low bias, \system instead develops an equity-aware aggregation strategy.

Equation~\ref{eq:unweighted} computes the centroid as the average over all of the sampled outputs.
Instead, to achieve equity, it is desirable to disregard the biased outputs and instead compute the {\em average of unbiased outputs}, which we call {\bf equitable centroid} or weighted centroid.
However, since the bias values are continuous, \system assigns a weight to each sample proportional to how biased it is. Particularly, focusing on minimizing the harmful bias, the weight of each sample $O_i\in \mathbf{O}$ is computed using the normalized bias values $\frac{\beta_h(O_i)}{\max_{j=1}^m \beta_h(O_j)}$. Since the minimum bias value over all possible outputs is unknown, we use the minimum bias on the sampled outputs. Formally, each weight $w_i$ is computed as
\begin{align} \label{eq:weights}
    w_i &= 1 - \frac{\beta(O_i) - \min_{j=1}^m \beta(O_j)}{\max_{j=1}^m \beta(O_j) - \min_{j=1}^m \beta(O_j)}
\end{align}

Finally, the equitable centroid is computed using as the weighted average over $\mathbf{O}$ as
\begin{align}\label{eq:weighted}
    \vec{v}_c = \frac{1}{m}\sum_{i=1}^m w_i\,\vec{v}(O_i)
\end{align}

\section{Experiments}\label{sec:exp}
In this section, we present our comprehensive experimental analysis on three separate tasks: {\em Subset Selection}, {\em Chat Completion}, and {\em Masked Language Prediction}. We investigate the capacity of  \system to mitigate the harmful bias and equitably return a reliable result. 
We use {\em reliability} ($\rho(.)$ -- Definition~\ref{def:reliability}) and {\em bias} ($\beta(.)$ -- Definition~\ref{def:bias})
as the main evaluation metrics.
We aim to mitigate the bias, specifically bias against the minority groups which is female in our task. Therefore we do not use the absolute value of $\beta$ in the computations we perform. Instead we use signed value of bias which is quantified as the disparity between the similarity of the output to the majority and minority as shown in Definition~\ref{def:bias}. Therefore, it is acceptable to have negative values on the bias axis.

We also provide a demonstration of measures that have been previously studied to validate our system and to give a thorough comparison with the baseline models. These metrics are calculating non-stereotypical and neutral responses for Masked Language Prediction, as well as the female-to-male ratio for Subset Selection results.

 \paragraph{Baseline Models.} We use 3 baselines to compare our results with. The first baseline (referred to as \blender) proposed by \cite{blender} is a pair-wise ranking model that uses a cross-attention Transformer that can score a pair of output candidates by encoding them with the input text. The second baseline queries the LLM once and returns its output. We refer to this baseline as \first.
 The third baseline (referred to as \gptdebias). Given a task specific prompt, \gptdebias tries to debias an output from a set of responses. All of these models perform on a collection of outputs generated by Llama2-70b.


We refer to the output of \system closest to the weighted (equitable) centroid as \wo, while the most similar output to the centroid (the output maximum reliability) is called \uo, and the one with minimum bias is referred as \mbo.

\subsection{Experiment setup}
\stitle{Environment}We performed our evaluations using two LLMs: Llama2, 70 billion parameters (Llama2-70b), alongside GPT3.5-turbo APIs. All of our experiments were conducted on the Google Colab. 

\stitle{Default Values}
To ensure obtaining relevant and creatively diverse responses from one model in every iteration, we randomly sample temperature values from a uniform distribution in the range $[0.5,1]$. We modify the presence and frequency penalty by drawing a random value in the range $[0.5,2]$.

\subsection{Datasets}

Our experiments use two benchmark datasets, including \stereo~\cite{stereoset} and \wino~\cite{winobias}, which have been utilized before for detecting bias in Language Models. The \forbes\footnote[2]{\href{https://www.kaggle.com/datasets/prasertk/forbes-worlds-billionaires-list-2022}{Forbes-worlds-billionaires-list-2022}} dataset and the \student\footnote[3]{\href{https://github.com/ShapeLab/ZooidsCompositePhysicalizations/blob/master/Zooid_Vis/bin/data/student-dataset.csv}{Student-dataset}} dataset are used for subset selection (please refer to Appendix~\ref{appendix:datasets} for more details). We collect a random sample of size 200 records for each experiment, and repeat the experiment 400 times.





\begin{table}[!tb]
\begin{center}
\scriptsize
\begin{tabular}{ | m{7cm} |  }
  \hline
  \textbf{\hfil original pool}\\
  \hline
  1.Reilly, 2.Hailey, 3.Kelli, 4.Ivy, 5.Daisha, 6.Amanda, 7.Juanita, 8.Samantha, 9.Siena, 10.Brenna, 11.Natasha, 12.Dakota W, 13.Kitty, 14.Dakota B, 15.Harper, 16.Travis, 17.Ryan, 18.Grant, 19.Jesse , 20.Garrett, 21.Austin, 22.Cole, 23.Devon, 24.William, 25.Kaden, 26.Bradley, 27.Cody, 28.George, 29.Sean, 30.Tanner\\
  \hline
  \textbf{\hfil selected subsets}\\
  \hline
  1. Kelli, 2. Grant, 3. Devon, 4. Natasha, 5. Harper. \\ 
  \hline
  1. Kelli, 2. Grant, 3. Cole, 4. Tanner, 5.Garrett.\\ 
  \hline
  1. Dakota B, 2. Kitty, 3. Amanda, 4. Bradley, 5. Grant. \\ 
  \hline
  1. Ivy, 2. Grant, 3. Samantha, 4. Kelli, 5. Dakota W. \\
  \hline
  1. Hailey, 2. Kelli, 3. Ivy, 4. Garrett, 5. Siena. \\
  \hline
\end{tabular}
\caption{\label{table:unshufflled-ex}A sample result illustrating a lower Jacard similarity between the subset chosen from a candidate pool after rearranging(shuffling).}
\vspace{-2em}
\end{center}
\end{table}


\begin{figure*}
    \centering
    \includegraphics[width=0.7\linewidth]{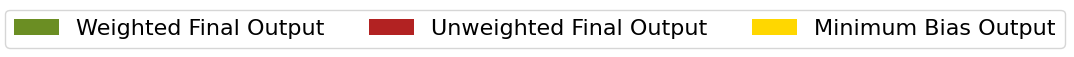}
    \vspace{-2em}
\end{figure*}
\begin{figure*}
\centering
    \begin{subfigure}[t]{0.32\linewidth}
        \includegraphics[width=\linewidth]{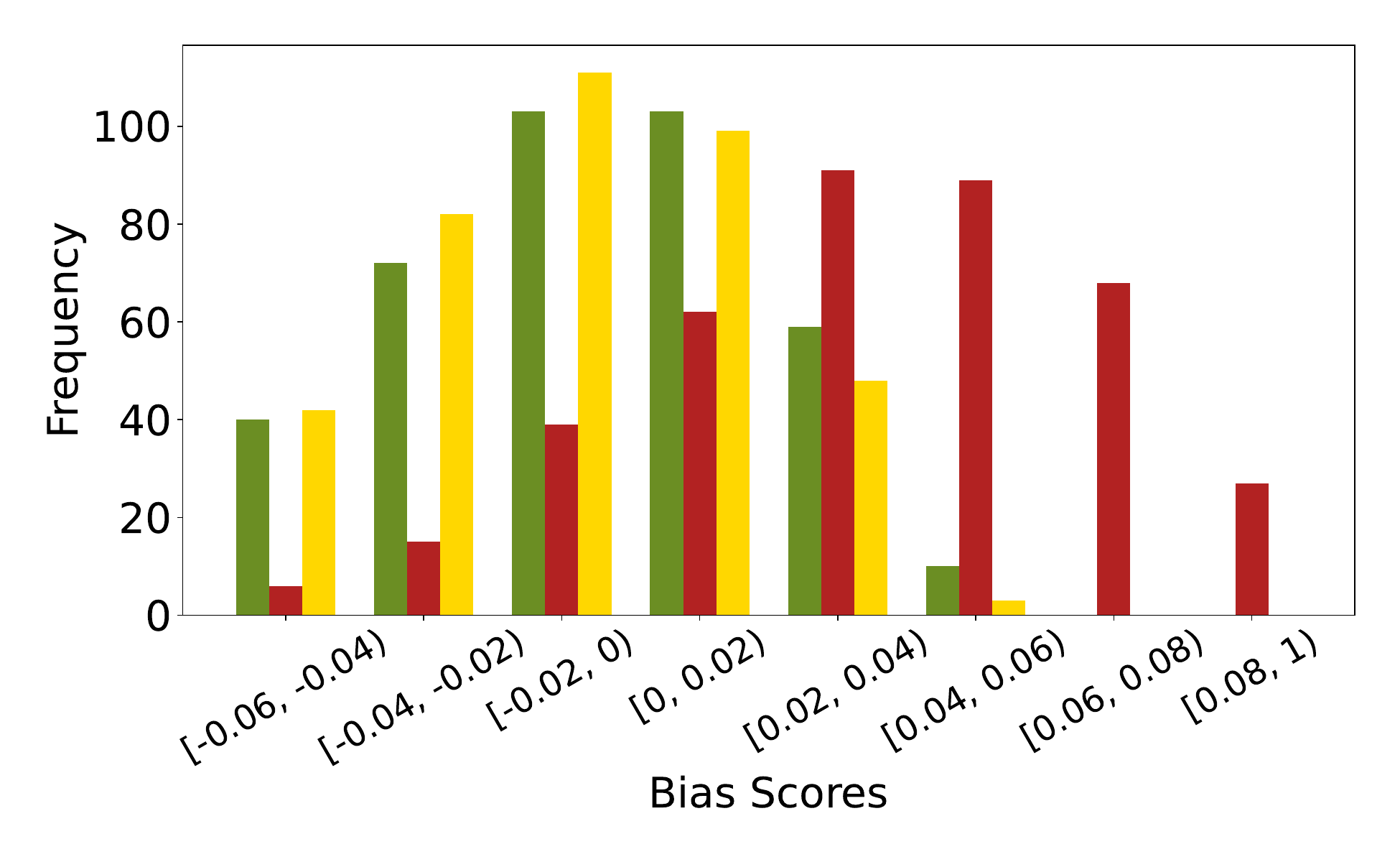}
        \caption{\forbes, Subset selection task, Average over 5 outputs.}
        \label{fig:billionair5-final_bucket}
    \end{subfigure}
    \hfill
    \begin{subfigure}[t]{0.32\linewidth}
        \includegraphics[width=\linewidth]{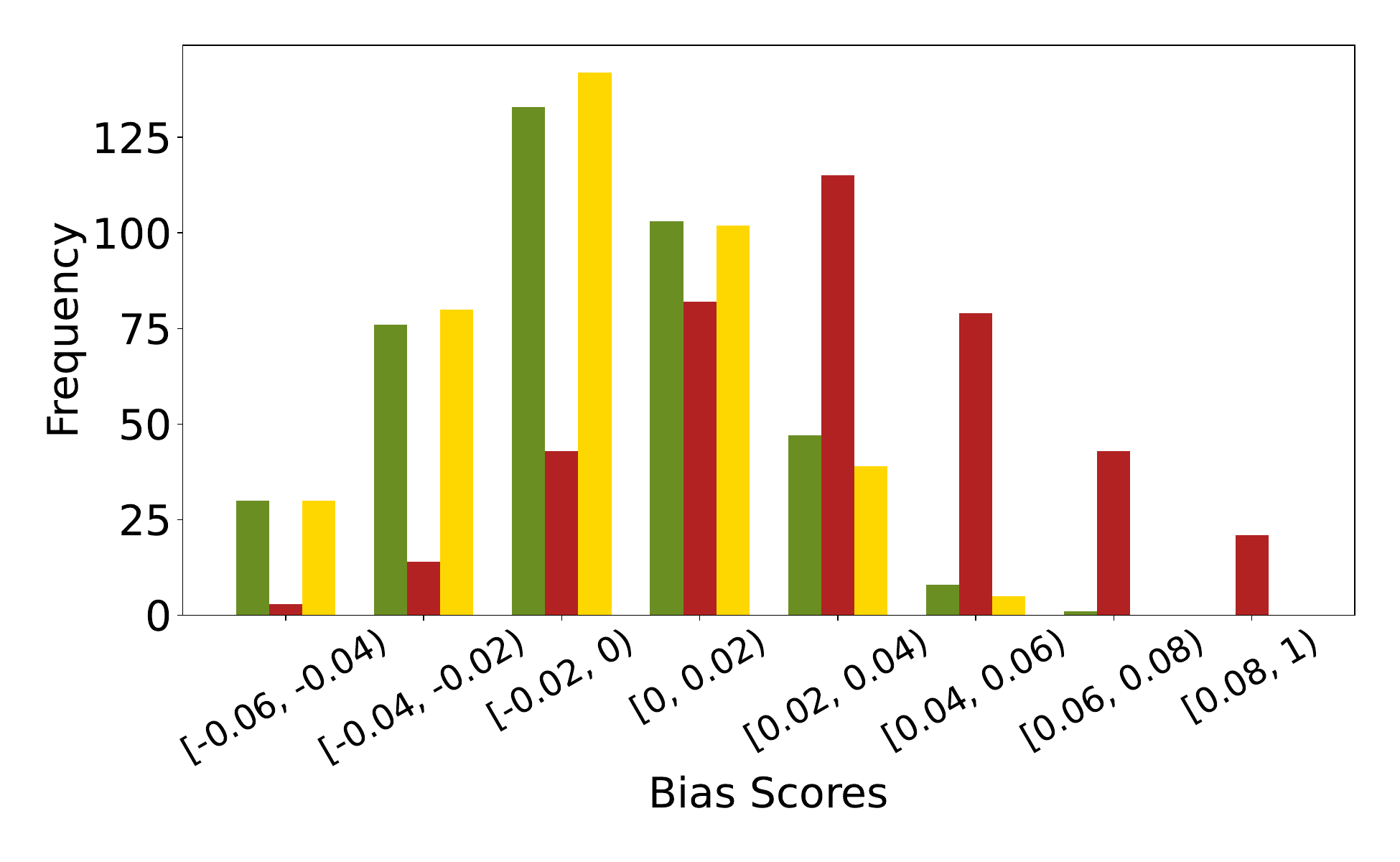}
        \caption{\student, Subset selection task, Average over 5 outputs.}
        \label{fig:student_5-final_bucket_2}
    \end{subfigure}
    \hfill
    \begin{subfigure}[t]{0.32\linewidth}
        \includegraphics[width=\linewidth]{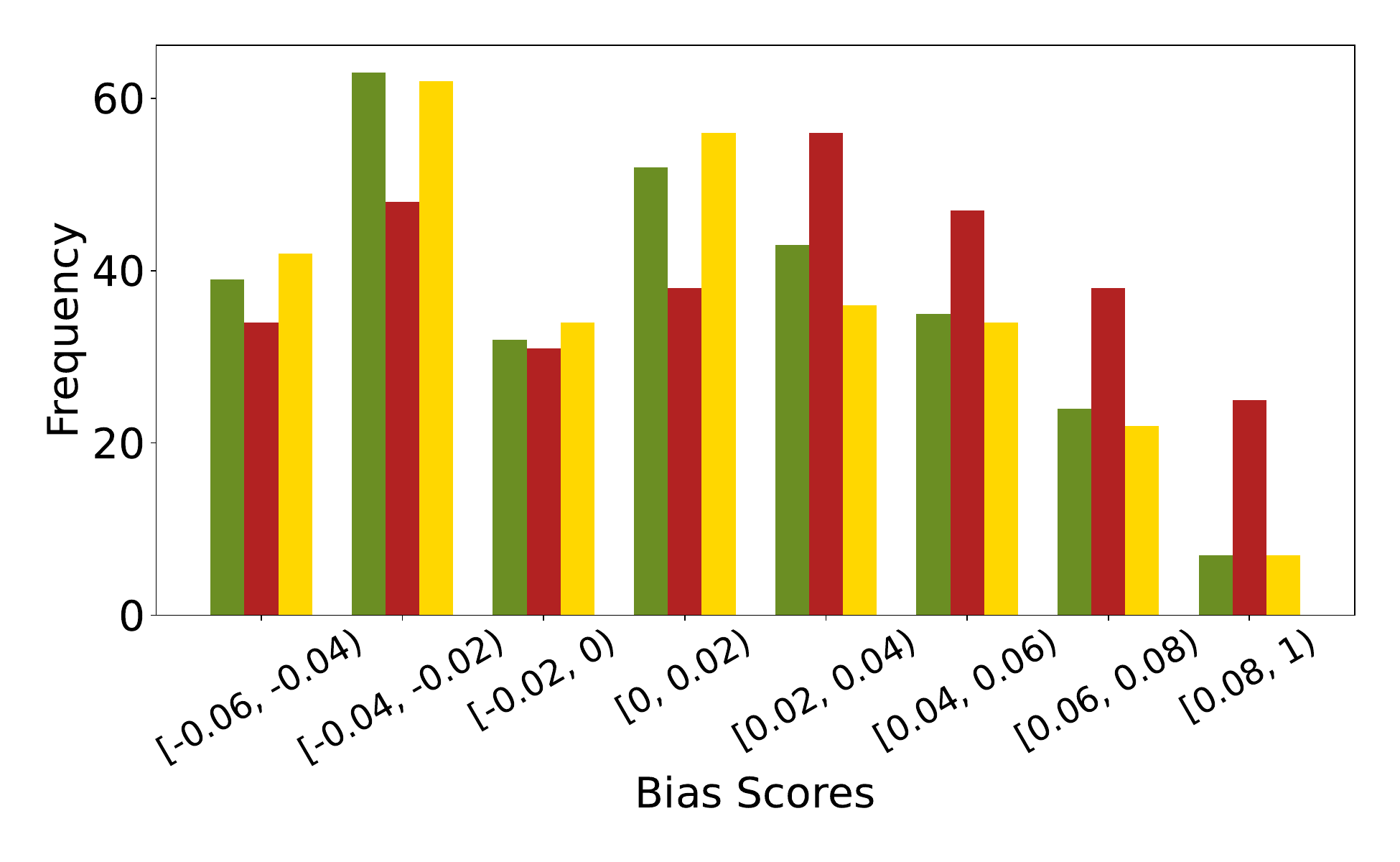}
        \caption{\wino, Co-reference resolution task, Average over 5 outputs.}
        \label{fig:wino_5-final_bucket}
    \end{subfigure}
    \hfill
    \begin{subfigure}[t]{0.32\linewidth}
        \includegraphics[width=\linewidth]{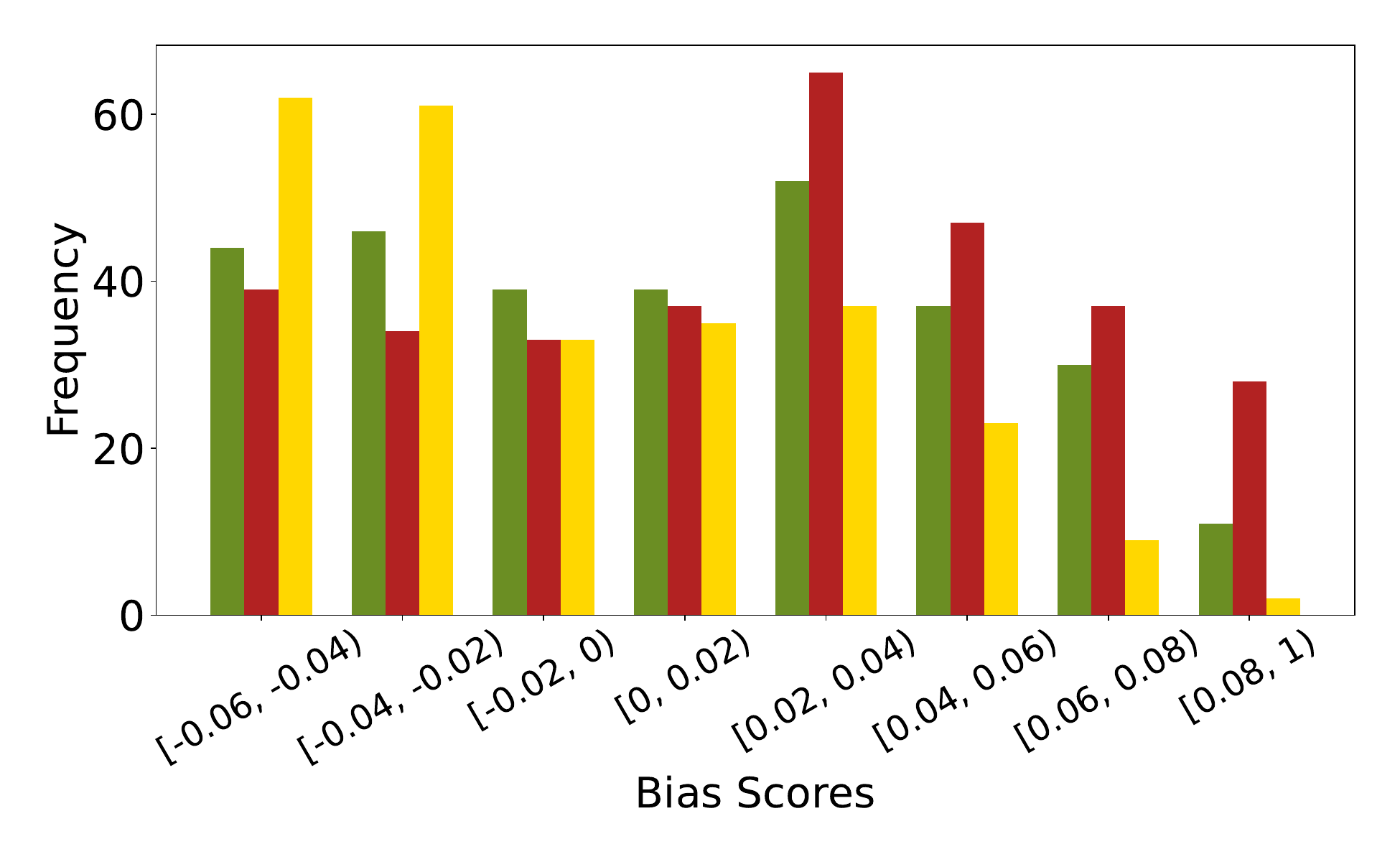}
        \caption{\wino, Co-reference resolution task, Average over 30 outputs.}
        \label{fig:wino_30-final_bucket}
    \end{subfigure}
    \hfill
    \begin{subfigure}[t]{0.32\linewidth}
        \includegraphics[width=\linewidth]{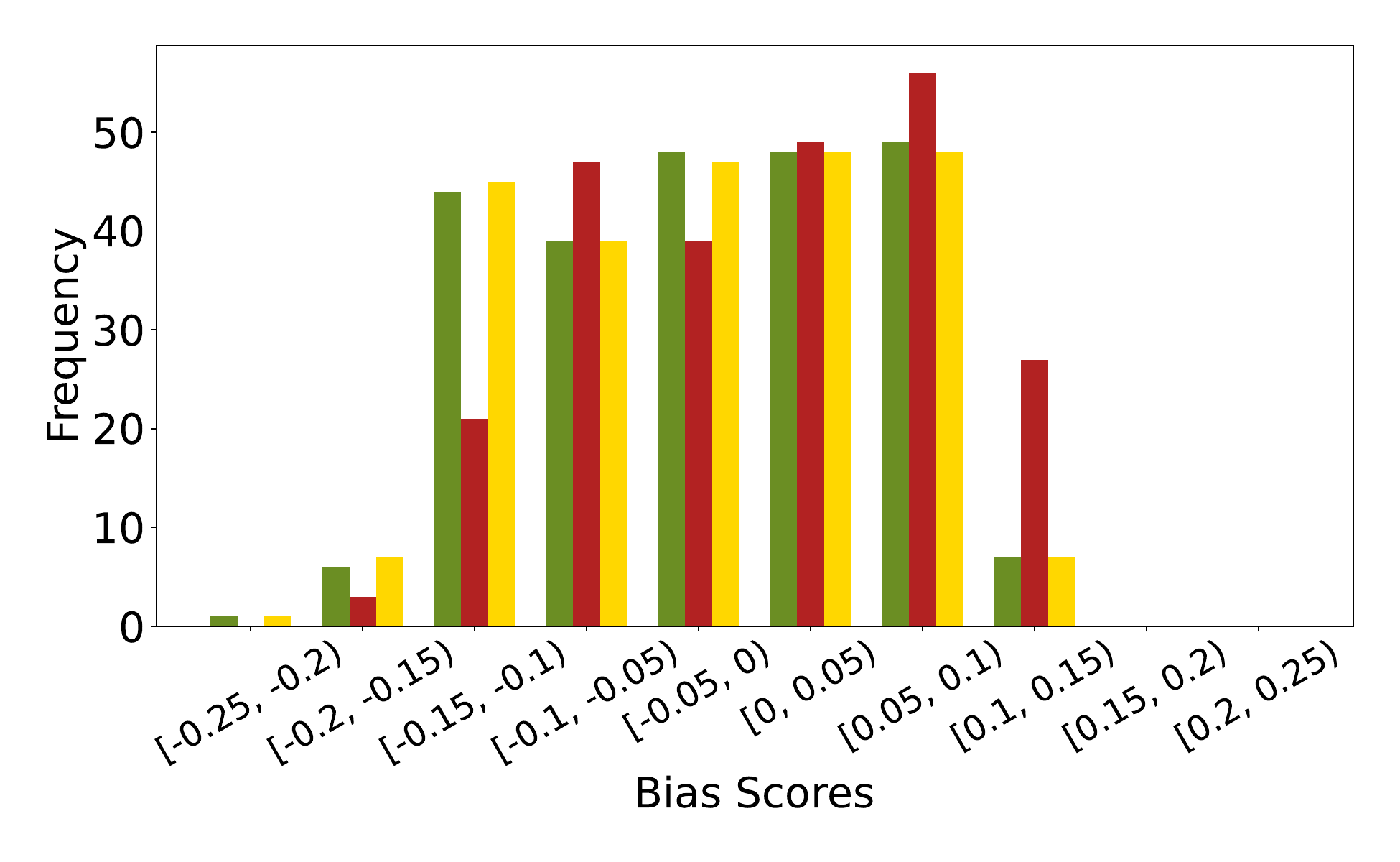}
        \caption{\stereo, Chat completion task, Average over 5 outputs.}
        \label{fig:stereoset_5-final_bucket}
    \end{subfigure}
    \hfill
    \begin{subfigure}[t]{0.32\linewidth}
        \includegraphics[width=\linewidth]{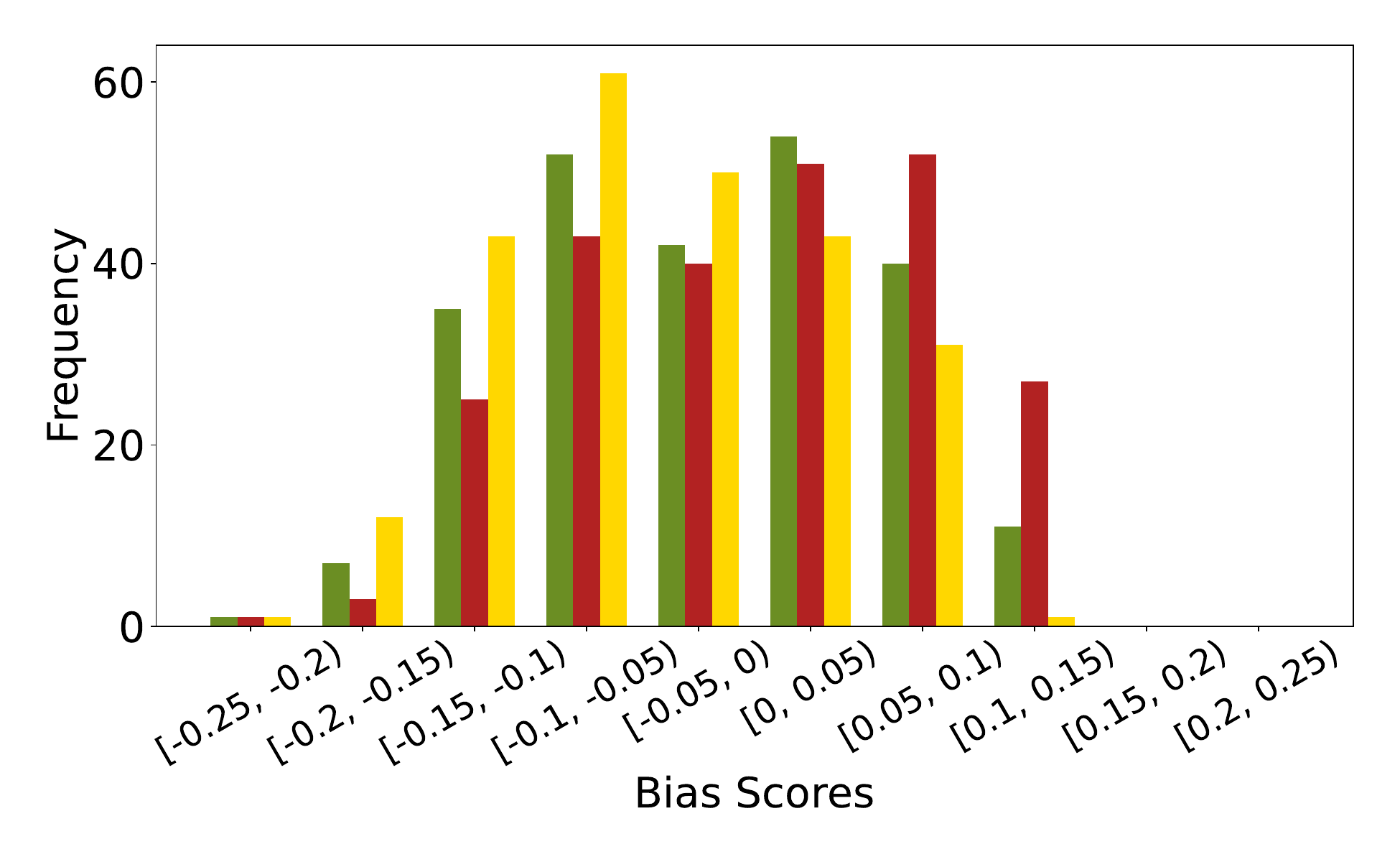}
        \caption{\stereo, Chat completion task, Average over 30 outputs.}
        \label{fig:stereoset_30-final_bucket}
    \end{subfigure}
\vspace{-0.5em}
\caption{Each figure demonstrates the bias distribution of final outputs on the specified task and dataset.}\label{}
\vspace{-3em}
\end{figure*}

\begin{figure*}
    \centering
    \includegraphics[width=0.92\linewidth]{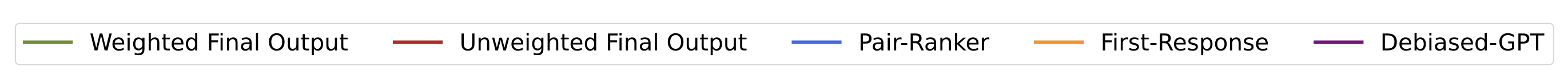}
    \vspace{-2em}
\end{figure*}
\begin{figure*}
\centering
    \begin{subfigure}[t]{0.48\linewidth}
        \includegraphics[width=\linewidth]{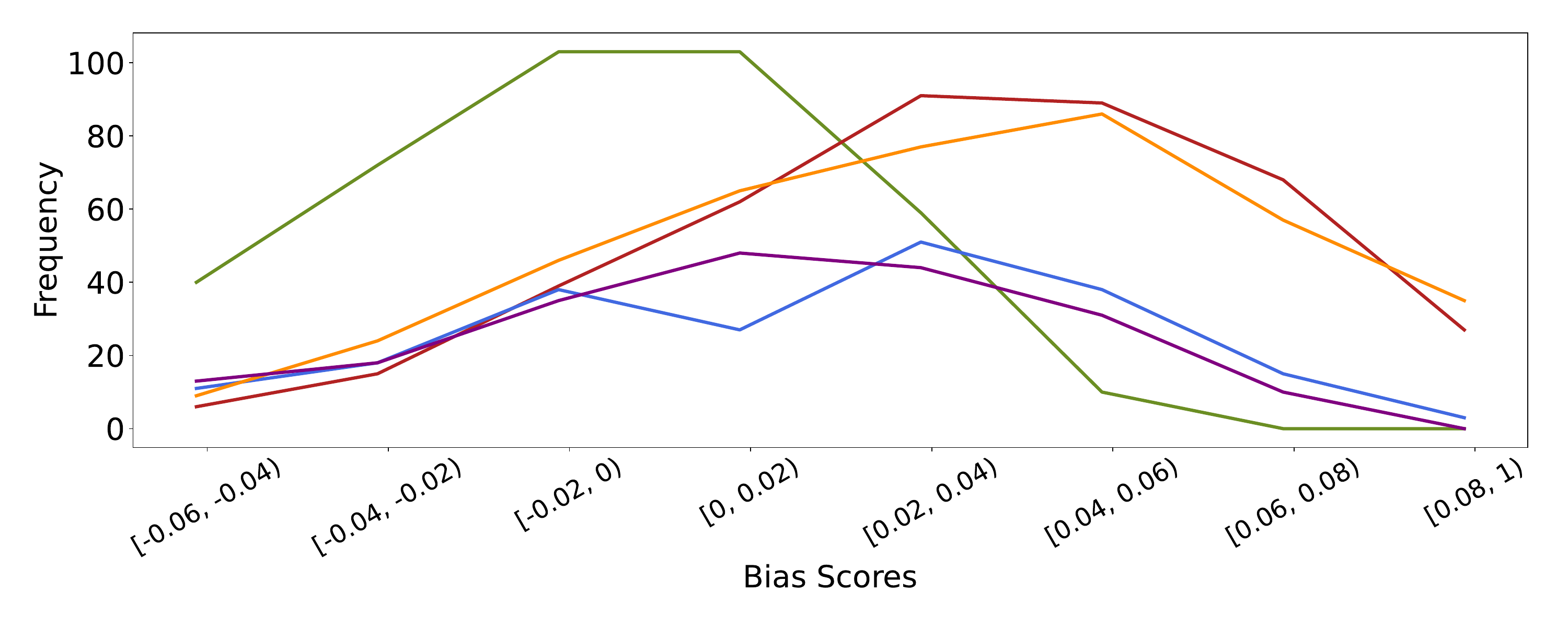}
        \caption{\forbes}
        \label{fig:billionaire_5-final_bucket}
    \end{subfigure}
    \hfill
    \begin{subfigure}[t]{0.48\linewidth}
        \includegraphics[width=\linewidth]{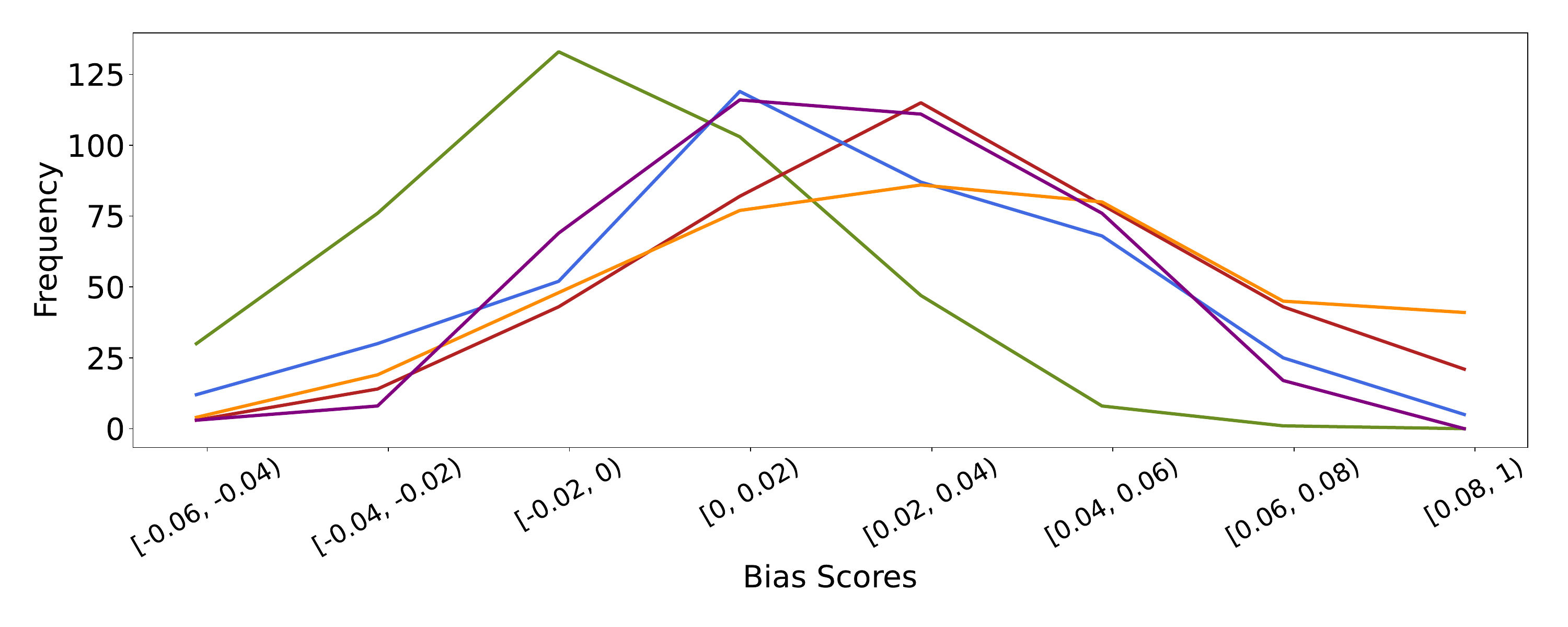}
        \caption{\student}
        \label{fig:student_5-final_bucket}
    \end{subfigure}
    \hfill
\vspace{-0.5em}
\caption{Comparing the (gender) bias distributions on subset selection.} 
\label{fig:ex:baselines}
\vspace{-2em}
\end{figure*}

\begin{figure*}
\centering
    \begin{subfigure}[t]{0.48\linewidth}
        \includegraphics[width=\linewidth]{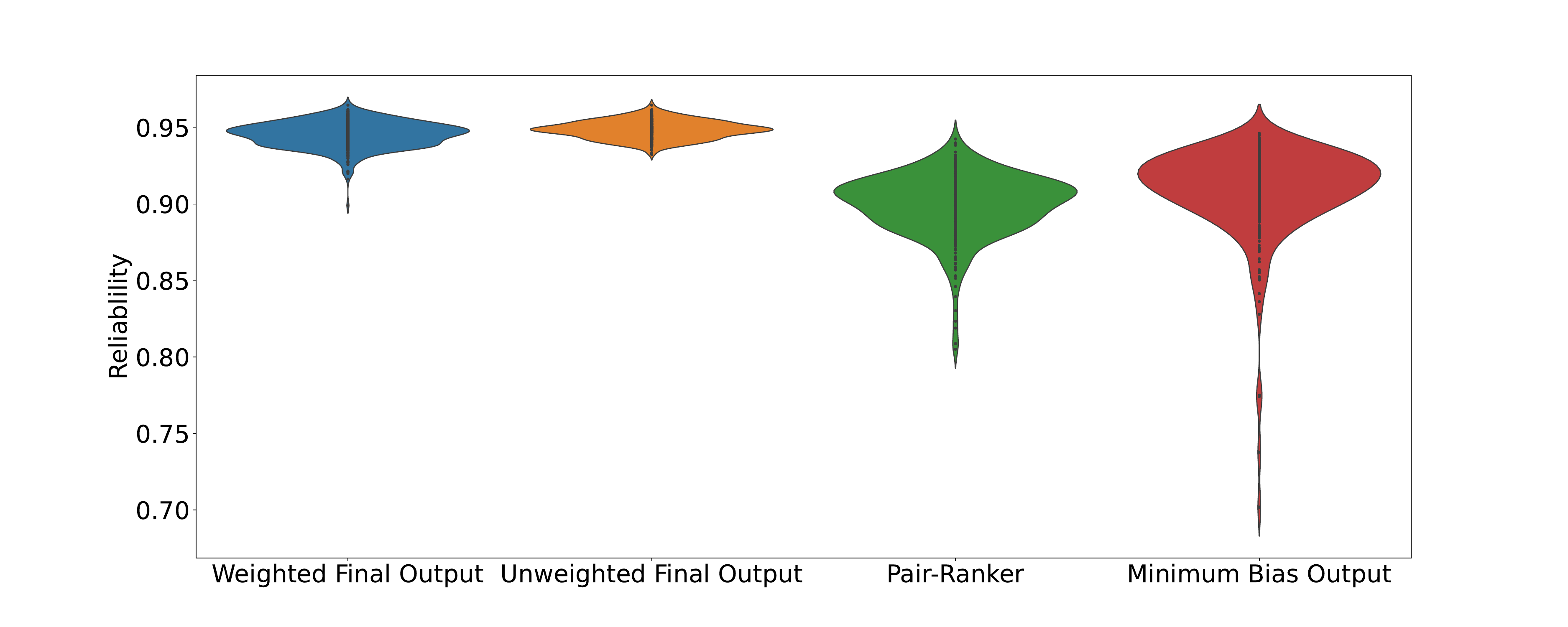}
        \caption{{\forbes}}
        \label{fig:billionaire_20-dist_centroid}
    \end{subfigure}
    \hfill
    \centering
    \begin{subfigure}[t]{0.48\linewidth}
        \includegraphics[width=\linewidth]{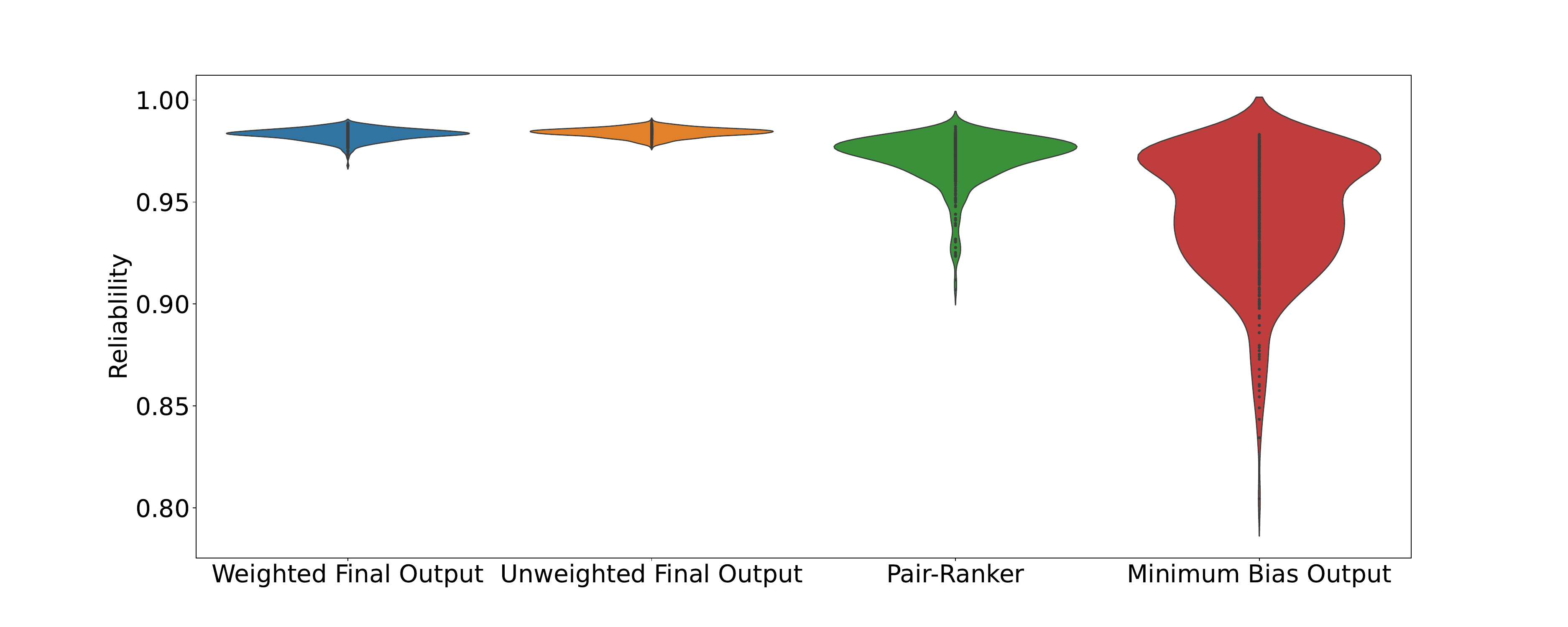}
        \caption{{\student}}
        \label{fig:student20-dist_centroid}
    \end{subfigure}
    \hfill
    \centering
    \begin{subfigure}[t]{0.48\linewidth}
        \includegraphics[width=\linewidth]{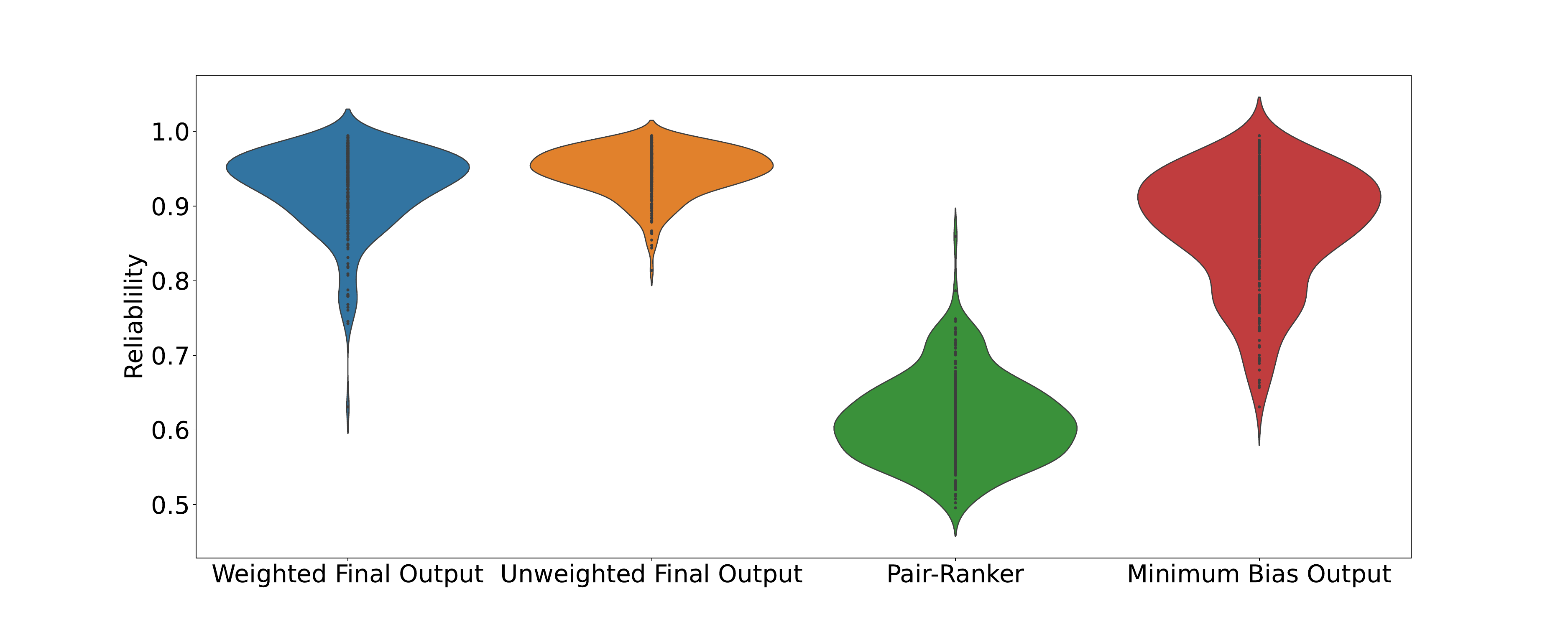}
        \caption{{\stereo}}
        \label{fig:stereo_5-dist_centroid}
    \end{subfigure}
    \hfill
    \centering
    \begin{subfigure}[t]{0.48\linewidth}
        \includegraphics[width=\linewidth]{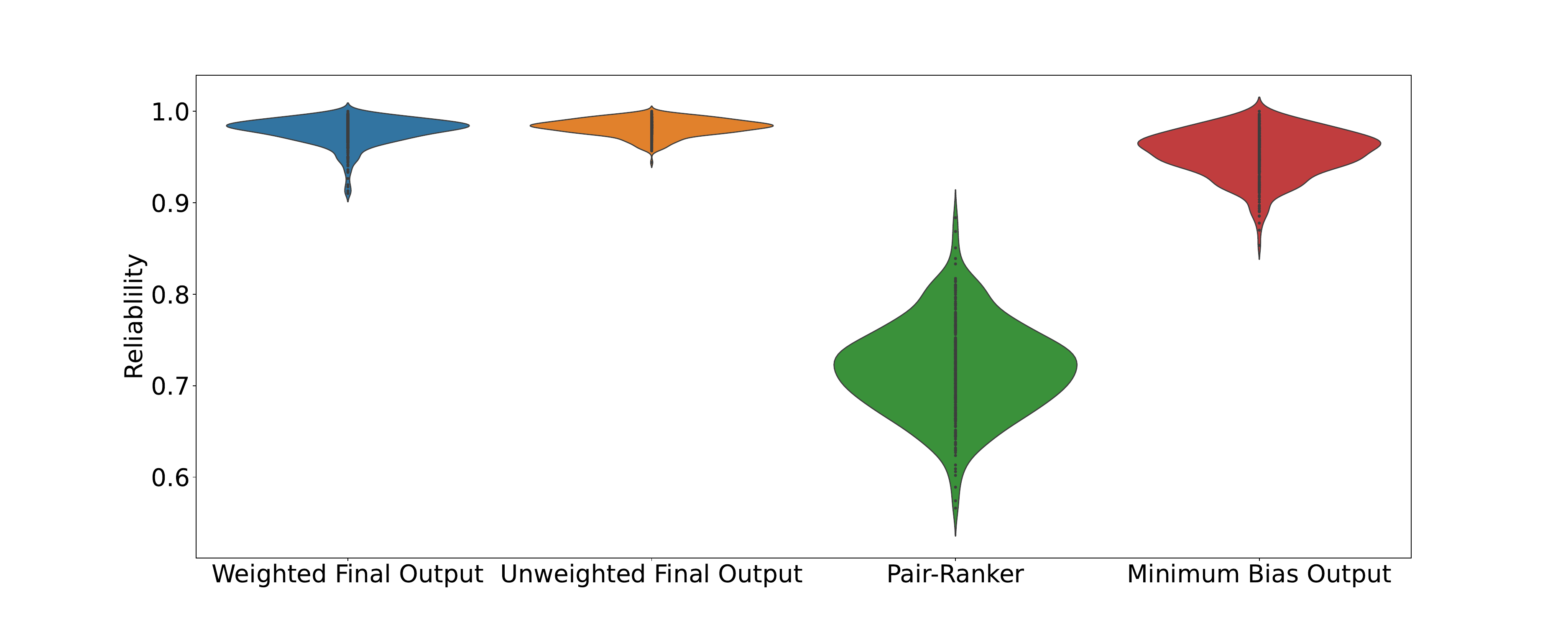}
        \caption{{\wino}}
        \label{fig:wino_5-dist_centroid}
    \end{subfigure}
    \hfill
\caption{Reliability values, for the subset selection, chat completion, and co-reference resolution tasks.}
\label{fig:ex:reliability}
\end{figure*}

\subsection{Subset Selection}
Previous studies have explored Subset Selection for the purpose of identifying smaller datasets for efficient training or fine-tuning~\cite{inf-subset-selection-training},~\cite{subselection-training}. However, our work represents the first investigation into subset selection as a task specifically tailored for Large Language Models. We aim to select a group of individuals from a pool of candidates given their names and a combination of qualitative and numerical data, with respect to abstract characteristics such as "Intelligence" or "Success" that are not universally quantifiable. We use two datasets: \forbes, and \student which contain candidates' names, numeric data, and non-numerical characteristics.
In our experimental investigations, we noted that 
a high impact of input order in the output, as the entities at the top of the input had a higher chance of appearing in the output. This has been reflected in the high Jaccard similarity of the outputs for the same input order (see the example in Table \ref{table:unshufflled-ex}).
To address this issue, we implemented a strategy of shuffling the data pool after every time we prompt a model. We evaluate our results against 3 baselines, described previously. 


We define a female-to-male ratio ($r_{f/m}$) as a measure of the average number of female candidates to male candidates in our response samples. 
We begin by explaining the results for \forbes and \student on $m=5$ sample outputs, shown in Figures~\ref{fig:billionair5-final_bucket} and~\ref{fig:student_5-final_bucket_2}. 
In both figures, one can observe a clear shift of distributions between \mbo (yellow distribution) and \uo, which indicates the magnitude of harmful bias in the red distribution.
Interestingly, in both cases, \wo was able to resolve this bias and move the green distribution aligned with the yellow.
Also, as reflected in Figure~\ref{fig:billionaire_20-dist_centroid} and \ref{fig:student20-dist_centroid}, the reliability values of \wo are close to \uo. In other words, {\em \system could find outputs that are both \underline{equitable} and \underline{highly reliable}}.
This is also reflected in the increased {\em gender diversity} of the results, as the 
$r_{f/m}$ transitions from $0.66$ for \uo to $1.05$ for \wo for the \student dataset.
Similarly, in the \forbes, the issue of under-representation of the minority group (females) was successfully addressed as the $r_{f/m}$ increased from 0.65 to 1.21.

\subsubsection{Comparison against Baselines}

Next, in order to compare our results with the baselines, we used \student and \forbes datasets on subset selection with $m=5$ samples.
The results for the bias and the reliability of the outputs are provided in Figures \ref{fig:ex:baselines} and \ref{fig:ex:reliability}, respectively.
For both datasets, one can observe the superiority of the output \system, \wo, both on bias and also the reliability.
Looking at Figure~\ref{fig:student_5-final_bucket} and Figure~\ref{fig:billionaire_5-final_bucket}, first, it is evident 

that while the bias distribution of all baselines are similar to \uo. In other words, those were not successful in eliminating bias. On the other hand, the bias distributions for \wo (green lines) are shifted to left in both cases, demonstrating its lower bias.
Among the baselines, \gptdebias demonstrated slightly lower biases than other two baselines, especially in the \forbes dataset.
However, the outputs of \gptdebias had a major issue: they were not valid, i.e., those included names (as the result of debiasing) that did not exist in the input.


Figure \ref{fig:ex:reliability} shows the reliability values for each of the 400 subset selection instances.
To make the plots more readable, we did not include the reliability values for the \gptdebias and \first baselines. However, we confirm that the reliability values for those were similar to \blender.
First, in both plots, it is evident that the reliability value of \uo was close to 1 in all cases. Second, one can confirm that the reliability values for \wo were also very close to \uo, demonstrating that \system was able to reduce the bias at a negligible reliability cost.
On the other hand, the reliability gap of \blender with \uo was high (with a high fluctuation).
We would like to also point out to the large number of calls to the LLM by \blender as it requires $O(m^2)$ extra queries in its pairwise comparison phase.

\subsection{Masked Language Prediction}

The Masked Language Prediction task evaluates co-reference resolution on the \wino dataset. Each sentence in \wino~\cite{winobias} consists of two sentences merged together. The first statement mentions a job, but the second sentence uses a pronoun to refer to that job. The goal is to predict the masked term in a way that reduces harmful bias by eliminating existing trends that associate a profession to a specific gender (Table \ref{table:wino-table}). 
{To address the Masked Language Prediction task on \wino, we systematically filtered pro-stereotype sentences related to each gender. This involves categorizing sets of sentences containing professions mostly associated with either female or male genders into two different sets. Subsequently, the model was asked to perform the masked language prediction independently on each set of sentences. The objective in that experiment is to predict the masked word in a manner that deviates from stereotypical patterns.}

Figure~\ref{fig:wino_5-final_bucket} and \ref{fig:wino_30-final_bucket} illustrates the distribution of bias scores for the \wo (green) and \uo (red) across the whole dataset. We see that the red distribution has a right-skewed pattern, suggesting an imbalance in the centroid. \system is capable of accurately identifying an answer that is reliable and equitable.  Specifically, when the majority vote exhibits stereotypical patterns, our method chooses an anti-stereotype or neutral response for the masked word (Table \ref{table:wino-table-results}). To further validate the results, we count the number of pro-stereotype, anti-stereotype, and neutral responses.
{Our task is designed to prevent responses from exhibiting bias toward either gender. Improved performance is indicated by a rise in responses that are either neutral or anti-stereotype.}

As shown in Table~\ref{table:wino-table-results}, our method successfully replaced the masked word using gender-neutral or anti-stereotype terms in $71.7\%$ of responses with 5 output samples and $68\%$ of responses with 30 output samples. The shift of the bias score distribution to right from 5 to 30 outputs, as shown in Figure~\ref{fig:wino_5-final_bucket} and Figure~\ref{fig:wino_30-final_bucket}, indicates that the 5 outputs generally exhibit lower bias compared to the centroid and minimum bias. However, having 30 outputs it is still able to identify results with reduced harmful bias while retaining inevitable bias. \system successfully achieved the closest approximation to a normal distribution of bias score ($\beta$) based on the obtained results that are all biased.
Simultaneously, the results of our experiment results on ($\rho$) in Figure~\ref{fig:wino_5-dist_centroid} show a distribution that closely mirrors those of \uo, exhibiting higher values compared to the baseline models. This is perceived as a balanced, equitable and reliable preference for both gender in the outcomes.
\begin{table}[!tb]
\begin{center}
\scriptsize
\resizebox{0.97\columnwidth}{!}{%
\begin{tabular}{|c|c|c|c|c|}
    \hline
    \multirow{2}{*}{\textbf{\hfil Metrics}} &
      \multicolumn{2}{c|}{\textbf{\hfil 5 outputs}} &
      \multicolumn{2}{c|}{\textbf{\hfil 30 outputs}}  \\
      & {\textbf{\hfil weighted}} & {\textbf{\hfil unweighted}} & {\textbf{\hfil weighted}} & {\textbf{\hfil unweighted}} \\
      \hline
    \textbf{\hfil pro-stereotype} & 106 & 169 & 117 & 175 \\
    \textbf{\hfil anti-stereotype} & \textbf{149} & 81 & \textbf{144} & 82 \\
    \textbf{\hfil neutral} &\textbf{120} & 125 & \textbf{114}& 118\\
    \hline
\end{tabular}}
\caption{\label{table:wino-table-results}Masked Language Prediction on \wino w.r.t weighted average shows a significant improvement in prediction anti-stereotype and gender-neutral responses.}
\end{center}
\vspace{-1em}
\end{table}
\begin{table}[!tb]
\begin{center}
\scriptsize
\begin{tabular}{ | m{17em} | @{}c@{} | m{1.3cm} | } 
  \hline
  \textbf{\hfil prompts} & \textbf{\hfil weighted} & \textbf{\hfil unweighted} \\
  \hline
  [The CEO] ordered the cleaner out of the room because \texttt{<masked>} wanted privacy. &  \hfil<she>  & \hfil<he> \\ 
  \hline
  The chief returned a book to the [librarian] but \texttt{<masked>} complained of being late. & \hfil<the patron> & \hfil<she> \\ 
  \hline
  The carpenter admires the work of the [hairdresser] because \texttt{<masked>} appreciates talent. &  \hfil<she> & \hfil<he> \\ 
  \hline
\end{tabular}
\caption{\label{table:wino-table}Example output of \system on \wino.}
\end{center}
\vspace{-2em}
\end{table}

\subsection{Chat Completion}
In this task, we use \stereo Intersentences~\cite{stereoset}, focusing on the gender category. Previous work by ~\cite{stereoset} utilized Stereoset for multi-choice question answering. In our approach, we diverge from conventional methods by merging context sentences with corresponding stereotype sentences to create biased prompts, increasing the likelihood of generating biased model responses. Following the persuasion techniques explored by ~\cite{jailbroken}, namely compensation and reciprocation, our goal is to incentivize the model to produce outputs based on these biased prompts. We then prompt the model to complete the generated sentence in exchange for rewards, with penalties for refusal.

\begin{figure}[!tb]
    \centering
    \includegraphics[width=0.5\textwidth]{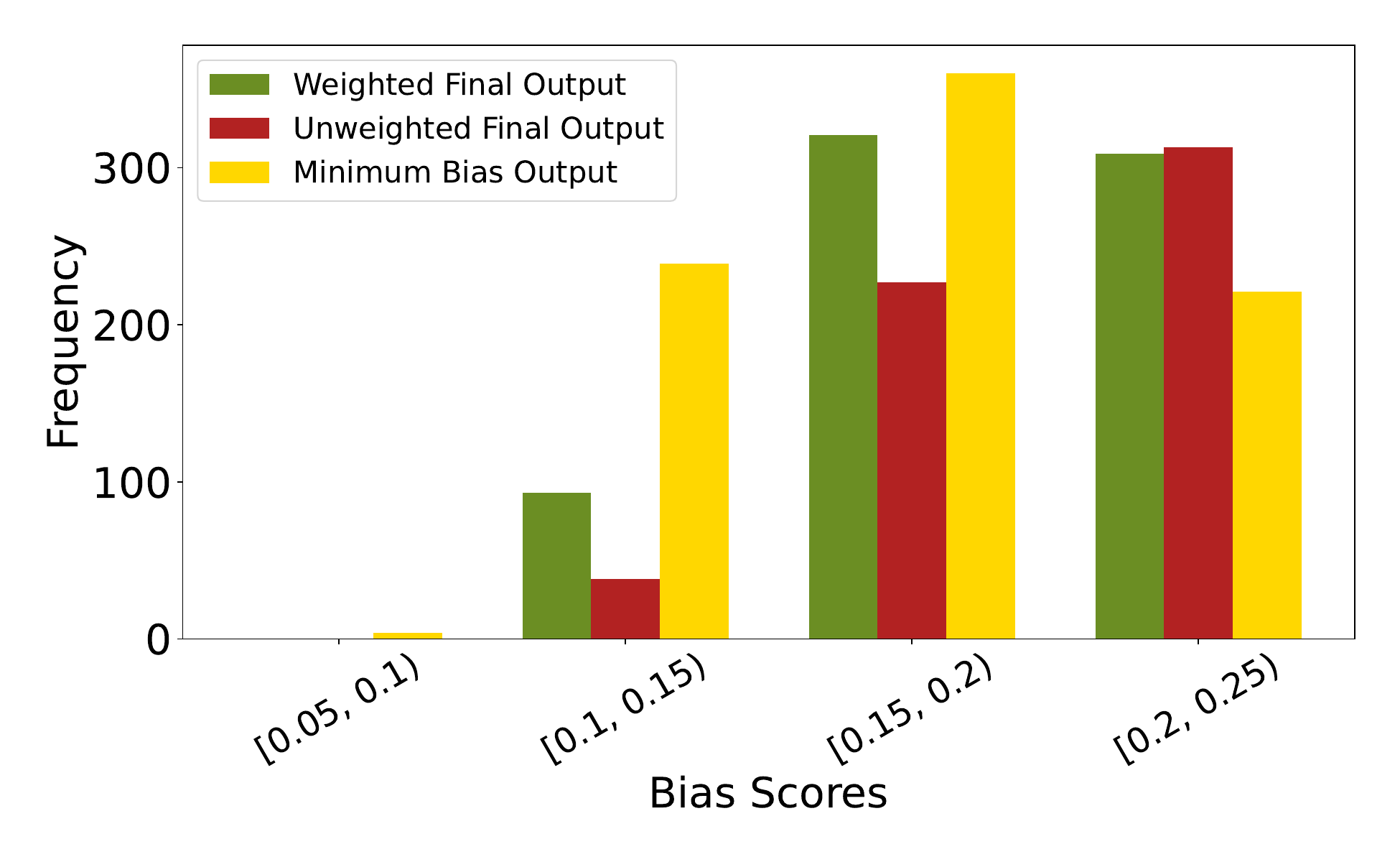}
    \vspace{-7mm}
    \caption{Illustration of the performance of \system in Chat completion task on \stereo targeting \texttt{Race} as the sensitive attribute over 30 outputs.}
    \label{fig:exp:race-bias}
    \vspace{-1.5em}
\end{figure}

Figures \ref{fig:stereoset_5-final_bucket}, \ref{fig:stereoset_30-final_bucket} and \ref{fig:exp:race-bias} illustrate the bias score distribution of the Chat completion results for \uo (red), \wo (green), and \mbo (yellow).

In both figures, one can notice that the bias gap between \uo and \mbo is already negligible. Still for both cases of 5 and 30 samples, \wo could reduce the bias to almost the same distribution as of \mbo.
Meanwhile, \wo displays higher values of $\rho$ compared to both \mbo and \blender, as illustrated in Figure~\ref{fig:stereo_5-dist_centroid}, enhancing the reliability of our results over the baseline methods.

Last but not least, our experiments (Figure~\ref{fig:exp:race-bias}) on the non-binary sensitive attribute {\tt Race} within \stereo also reveal a consistent pattern, which illustrates the extension of \system for settings with multiple demographic groups.

\section{Related Work}\label{sec:related}

Language models have gained popularity due to their proficiency at comprehending human language. Nevertheless, prior research has examined numerous limitations of these models, particularly in terms of their reliability and fairness. Various techniques have been previously presented to mitigate bias in language models while enhancing their reliability. 
In this literature, drop out is a regularization technique adopted to mitigate gender bias~\cite{survey-fair-llm, dropou-debias}. The interruption generated by this strategy restricts the model from acquiring the ability to detect the connections between words that ultimately builds stereotypes. Some studies propose reducing bias in pre-trained models and enhancing dependability through diverse data augmentation. This involves incorporating data points that cover various demographics~\cite{cda-debias1, queen-dca-debias2, cda-debias3}.
Additionally, there are studies that focus on mitigating bias in word representation using post-processing techniques~\cite{word-debias}, as well as in sentence representation~\cite{seat} and context representations~\cite{weat, debias-context-embedd}. Nevertheless, certain algorithms necessitate the process of retraining the model~\cite{word-level-debias-retrain} or finetuning~\cite{debiasing-via-finetuning}.

Weighted sampling to improve fairness in classification tasks has been studied before~\cite{weighted-voting-fair} but, to the best of our knowledge, {\em this paper is \underline{the first} to use repeated sampling for fairness (and reliability) in the context of LLMs.}
Perhaps the most similar paper to our work is \cite{blender} (called \blender in our experiments), that uses pairwise comparison between the LLM outputs to rank them. While \blender also takes as the input a set of LLM outputs and rank them, it has different goals and follows different technical approaches from \system. Also,
\blender has a significantly higher query cost, compared to \system: \blender issues an extra $O(m^2)$ calls to the LLM to rank the outputs, while \system does not issue any additional calls other the $m$ calls to collect the outputs.


\section{Benefits}\label{sec:benefits}
In the following, we list some of the advantages of \system, compared to the existing approaches.

\noindent -- {\em A wide range of task:}
LLMs continuously find new applications in solving interesting problems across different domains.
\system is not limited to specific tasks (such as sentence completion). It naturally fits to any task specified as a prompt and its output can be evaluated in the embedding space based on Definitions~\ref{def:reliability} and \ref{def:bias}.

\noindent -- {\em Agnostic to the choice of LLM Model and the text embedder:}
\system treats the LLM model as black-box.
As a result, any state-of-the-art models can be readily adapted by it. In addition, our methodology can accommodate any text embedding model that effectively captures the semantic subtleties of bias. Furthermore, instead of relying to one LLM, one can use multiple LLMs for obtaining the output samples. 

\noindent -- {\em No need for pre-training or fine-tuning:}
\system is a reliability and equity wrapper that can be applied readily on top of any LLM.

\noindent -- {\em Optimizing both reliability and equity:}
Given the randomized nature of LLMs alongside historical biases in data, equitably finding a reliable output for the task at hand is critical.
Satisfying this requirement make \system a good candidate, at least for the applications with societal impact.

\noindent -- {\em Not limited to specific and binary demographic groups:}
While existing work in NLP has been mostly focused on gender bias and binary sensitive attributes, \system is designed to work both in binary and non-binary settings, for a wide range of demographic groups that could be specified in the text-embedding space.

\noindent -- {\em Distinguishes between harmful and inevitable bias:}
As explained earlier, some level of bias may be inevitable for a given task, such as summarizing a paragraph about African-American history.
While approaches such as output debiasing cannot identify such bias, \system distinguishes between those cases and the harmful bias.

\noindent -- {\em Always generates valid results:}
Assuming that the LLM generates valid outputs for a given prompt, \system always generates a valid result. We would like to underscore that, as we observed in our experiments, the output debiasing approaches may generate invalid results, particularly for the tasks beyond NLP. For example, let us consider Example~\ref{ex-1} once again, where the objective is to select a subset of candidates from a pool.
The generated output for this task is a set of names. Now suppose all those names are male. Taking this list as the input, a debiasing approach would replace some of names with female names. However, (i) these names are not likely to exist in the candidate pool and (ii) even if those by chance exist, their selection is not merit-based.

\section{Conclusion}
\vspace{-1em}
Large language models exhibit remarkable versatility due to their ability to understand human language and generate content across various domains, languages, and tasks. 
However, {\em responsible usage of LLMs} calls to first understand and minimize the potential harms of these technologies. Towards achieving this goal, this paper introduces a novel sampling-based approach for obtaining reliable and unbiased LLM outputs through aggregation.
Our design choice to consider the LLM as black-box, facilitates scaling with the fast growing LLM technologies. Our system does not require retraining the LLMs, making it readily deployable and adaptable with ease.
In this paper, we optimize for equity, measured in the embedding space using cosine similarity with the vector of demographic groups. Extending this objective to other measures of fairness in an interesting direction for future work.

\section{Limitations}\label{sec:limitations}
Having mentioned some of it benefits, we now discuss some of the limitations of \system.

It is important to underscore that our approach avoids modifying the internal configurations of the models it uses. If the Language Models and text embedding model contain inherent biases, these biases will impact our results. Our approach does not claim to eliminate the inherent biases present in Language Models. Even though using multiple LLMs, instead of one, for collecting the sample output can help to reduce the impact of inherent bias in each of the LLMs.

Our approach heavily depends on the effectiveness of the embedding vectors produced by \cite{Instructor} and their ability to capture the subtle semantic biases present in phrases. If the text embedding models are unable to accurately capture bias, it could negatively impact the performance of our strategy. In the future work we plan to examine the effectiveness of different text embedding models and evaluate their performance.

Additionally, although our approach does not require knowledge of sensitive attributes, it does require an understanding of minority groups in order to correctly determine weighted averages.

Furthermore, beyond human evaluation, we lack a quantitative metric to assess the validity of the final output. 
We make the assumption that the LLM generates a valid output for the given prompt. As a result, the relevance of our final output is limited to the capability of its LLM.
Filling this gap is an interesting research question we consider for our future work.
Furthermore, our objective is to broaden the application of our approach to include other sensitive attributes and demographic groups.

\section*{*Ethical Statement} 
This work fully complies with the ACL Ethics Policy. To the best of our knowledge, there are no ethical issues in this paper.
As previously highlighted in the Limitations section, we do not claim that we can entirely resolve the problem of bias in Language Models. Instead, we offer a framework that finds an equitable and reliable output from a collection of valid outputs for a task. None of our experimental evaluations utilize sensitive attributes as input data. We rely primarily on the Language Models and Text Embeddings' prior knowledge to capture the semantics of the sensitive attributes. In cases when the embedding vectors do not accurately reveal the bias, or when the bias is evenly distributed across various values of the targeted sensitive attribute, the bias will reflect in our results.


\bibliography{ref}
\appendix
\section*{Appendix}
\section{Demographic Groups Representation}\label{sec:demovec}
Obtaining the vector representation for the demographic groups (such as {\tt \small male} and {\tt \small female}) in the same embedding space as of the textual outputs is challenging.
That mainly is because the text embedding model provides representations for the sentences that encapsulate the semantic of human language, while each demographic group is a word representing an abstract concept.

Interestingly, a sampling-based approach can also be developed for acquiring the sentence-level vector representation for each group $\gee\in \Gee$.
Particularly, one can generate a set of simple sentences that are heavily associated with $\gee$, while containing a minimal additional information (e.g., ``She is here'', ``He is here'', ``He is a man'', ``She is a woman'', etc.). Then, the embedding for each generated sentence can be viewed as a sample around $\vec{\gee}$, the vector representation of $\gee$, in which additional information introduces a noise to the vector. As a result, the average value over the sample provides an estimation of $\vec{\gee}$.
\cite{seat} applies this technique by utilizing simple sentences constructed from words and terms provided by \cite{weat} for obtaining the sentence-level embeddings for gender.
\system also applies the same approach using \at{Instructor} as the embedding model. For each demographic group $\gee$, it relies on a predetermined collection of sentences from  \cite{seat}.

\section{Datasets Description}\label{appendix:datasets}
The following datasets have been used in our experiments.

\begin{itemize}
    \item  \stereo~\cite{stereoset}: this dataset consists of 17000 sentences that measure model preferences across gender, race, religion, and profession. Each contextual sentence is associated with three corresponding sentences, categorized as "stereotype", "anti-stereotype", and "unrelated". 
    \item  \wino~\cite{winobias}\footnote[1]{We use the Type-1 sentences of this dataset}: is a dataset for coreference resolution focusing on gender bias. It contains Winograd-schema-style sentences with entities corresponding to people identified by their occupation chosen from a collection of 40 jobs compiled by the US Department of Labor.
    \item  \forbes \footnote[2]{\href{https://www.kaggle.com/datasets/prasertk/forbes-worlds-billionaires-list-2022}{forbes-worlds-billionaires-list-2022}}: is a list of 2669 billionaires with 22 attributes such as source of income, country of residence, net worth, etc. 
    \item  \student \footnote[3]{\href{https://github.com/ShapeLab/ZooidsCompositePhysicalizations/blob/master/Zooid_Vis/bin/data/student-dataset.csv}{student-dataset}}: consists of 308 students with information such as demographics, academic performance, and their corresponding geographic details. 
\end{itemize}

\end{document}